\documentclass[letterpaper, 10pt, conference]{ieeeconf} 

\IEEEoverridecommandlockouts 

\newcommand{\site}{\textbf{\url{https://gamma.umd.edu/meteor}}}
\newcommand{\rain}{\textsc{Meteor}\xspace}

\makeatletter
\newcommand\footnoteref[1]{\protected@xdef\@thefnmark{\ref{#1}}\@footnotemark}
\makeatother

\newcommand{\shorteq}{%
  \settowidth{\@tempdima}{-}
  \resizebox{\@tempdima}{\height}{=}%
}

\usepackage[colorlinks=true, citecolor=magenta]{hyperref}
\usepackage[table]{xcolor}

\usepackage{amssymb,fge}

\usepackage{amsmath, amssymb}
\usepackage{pifont}
\newcommand{\cmark}{\ding{51}}%
\newcommand{\xmark}{\ding{55}}%
\usepackage{subcaption}
\usepackage[utf8]{inputenc}
\usepackage[english]{babel}
\usepackage[linesnumbered,ruled,vlined]{algorithm2e}
\usepackage[font=small,belowskip=-1.5pt]{caption} 
\usepackage{anyfontsize}

\usepackage{array}
\usepackage{graphicx}
\usepackage{amsfonts}
\usepackage[11pt]{moresize}
\usepackage{bm}
\usepackage{soul}
\usepackage{hhline}
\usepackage{multirow, makecell, multicol}
\usepackage{float}
\usepackage{booktabs,wrapfig}

\usepackage{amsthm}
\usepackage{color}
\usepackage{transparent}
\usepackage{url}
\usepackage{footmisc}
\usepackage{setspace}
\usepackage{mathtools}
\usepackage{threeparttable}
\usepackage{gensymb}
\usepackage[export]{adjustbox}

\theoremstyle{plain}

\title{\rain: A Dense, Heterogeneous, and Unstructured Traffic Dataset With Rare Behaviors}

\author{Rohan Chandra$^{1*}$, Xijun Wang$^{1*}$,  Mridul Mahajan$^{2}$, Rahul Kala$^{2}$, Rishitha Palugulla$^{3}$,\\ Chandrababu Naidu$^{3}$, Alok Jain$^{3}$, and Dinesh Manocha$^{1,4}$\\
{\small Dataset, Code, and Video at \site}
\thanks{$^{*}$Denotes equal contribution.}
\thanks{$^{1}$Department of Computer Science, University of Maryland, College Park, USA.
{Corresponding email: \tt\small rchandr1@umd.edu}}%
\thanks{$^{2}$ Centre for Intelligent Robotics, Indian Institute of Information Technology, Allahabad, India.
}%
\thanks{$^{3}$ NavAjna Technologies Pvt. Ltd.}
\thanks{$^{4}$Department of Electrical and Computer Engineering, University of Maryland, College Park, USA.}
}

\begin{document}

\maketitle
\thispagestyle{empty}
\pagestyle{empty}


\begin{abstract}

We present a new traffic dataset, \rain, which captures traffic patterns and multi-agent driving behaviors in unstructured scenarios. \rain~consists of more than $1000$ one-minute videos, over $2$ million annotated frames with bounding boxes and GPS trajectories for $16$ unique agent categories, and more than $13$ million bounding boxes for traffic agents. \rain is a dataset for rare and interesting, multi-agent driving behaviors that are grouped into traffic violations, atypical interactions, and  diverse scenarios. Every video in \rain is tagged using a diverse range of factors corresponding to weather, time of the day, road conditions, and traffic density. We use \rain to benchmark perception methods for object detection and multi-agent behavior prediction. Our key finding is that state-of-the-art models for object detection and behavior prediction, which otherwise succeed on existing datasets such as Waymo, fail on the \rain dataset. \rain marks the first step towards the development of more sophisticated perception models for dense, heterogeneous, and unstructured scenarios.
\end{abstract}
\section{Introduction}
\label{sec: introduction}

Recent research in learning-based techniques for robotics, computer vision, and autonomous driving has been driven by the availability of datasets and benchmarks. Several traffic datasets have been collected from different parts of the world to stimulate research in autonomous driving, driver assistants, and intelligent traffic systems. These datasets correspond to highway or urban traffic, and are widely used in the development and evaluation of new methods for perception~\cite{chandra2019roadtrack}, prediction~\cite{social-lstm}, behavior analysis~\cite{chandra2020stylepredict}, and navigation~\cite{ppp}.

Many initial autonomous driving datasets were motivated by computer vision or perception tasks such as object  recognition, semantic segmentation or 3D scene understanding. Recently, many other datasets have been released that consist of point-cloud representations of objects captured using LiDAR, pose information, 3D track information, stereo imagery or detailed map information for applications related to 3D object recognition and motion forecasting.
Many large-scale motion forecasting datasets such as Argoverse~\cite{chang2019argoverse}, and Waymo Open Motion Dataset~\cite{waymo}, among others, have been used extensively by researchers and engineers to develop robust prediction models that can forecast vehicle trajectories. However, existing datasets do not capture the rare behaviors or heterogeneous patterns.  Therefore, prediction models trained on these existing datasets are not very robust in terms of handling challenging traffic scenarios that arise in the real world.
 \begin{figure*}[t]
    \centering
    \includegraphics[width=\textwidth]{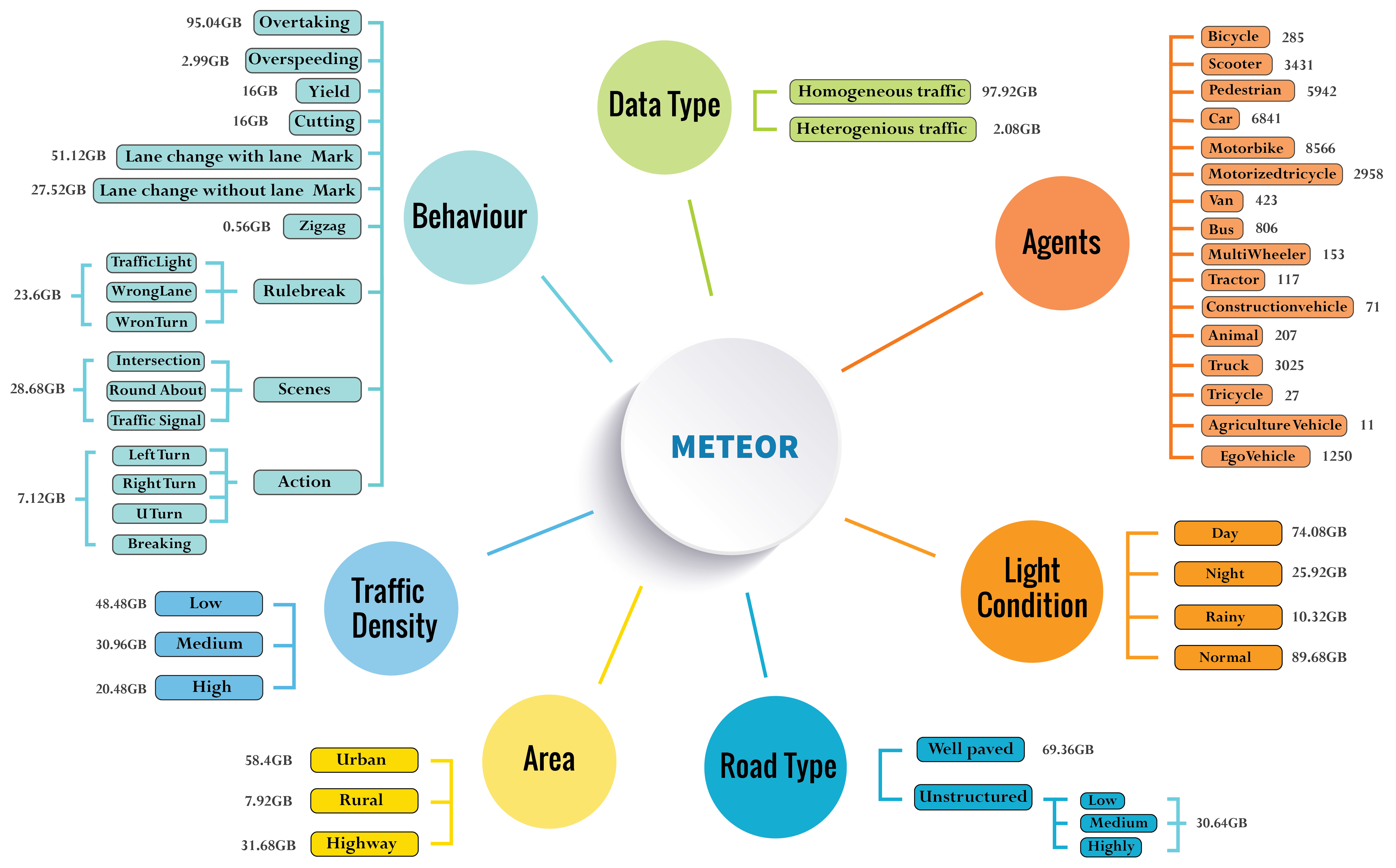}
    \vspace*{-0.12in}
    \caption{\textbf{\rain:} We summarize various characteristics of our dataset in terms of scene: traffic density, road type, lighting conditions, agents (we indicate the total count of each agent across $1250$ videos), and behaviors, along with their size distribution (in GB). The total size of the current version of the dataset is around $100$GB, and it will continue to expand. Our dataset can be used to evaluate the performance of current and new methods for perception, prediction, behavior analysis, and navigation based on some or all of these characteristics. Details of the organization of our dataset are given at \site. }
    \label{fig: infographic}
    \vspace{-10pt}
\end{figure*}

A major challenge currently faced by research in autonomous driving is the \textit{heavy tail problem}~\cite{chang2019argoverse, waymo}, which refers to the challenge of dealing with rare and interesting instances. 
There are several ways in which existing datasets currently address the heavy tail problem: 
\begin{enumerate}
    \item \textbf{Mining:} The Argoverse and Waymo datasets use a mining procedure that includes scoring each trajectory based on its ``interestingness'' to explicitly search for difficult and unusual scenarios~\cite{chang2019argoverse, waymo}. 
    \item \textbf{Diversifying the taxonomy:} Train the prediction and forecasting models to identify the unknown agents at the time of testing. This approach necessitates annotating a diverse taxonomy of class labels. Argoverse and nuScenes~\cite{nuscenes2019} contain $15$ and $23$ classes, respectively. 
    \item \textbf{Increasing dataset size:} This approach is to simply collect more data with the  premise that collecting more traffic data will likely also increase the number of such scenarios in the dataset.
\end{enumerate}
In spite of many efforts along these lines, existing datasets manage to collect only a handful of such instances, due to the infrequent nature of their occurrence. For example, the Waymo Open Motion dataset~\cite{waymo} contains only atypical interactions and diverse scenarios while the Argoverse dataset~\cite{chang2019argoverse} contains only atypical interactions. There is clearly a need for a different approach to addressing the heavy tail problem. Our solution is to build a traffic dataset from videos collected in India, where the inherent nature of the traffic is dense, heterogeneous, and unstructured. The traffic patterns and surrounding environment in parts of India are more challenging.
than those in other parts of the world. This includes high congestion and traffic density. Some of these roads are unmarked or unpaved. Moreover, the traffic agents moving on these roads correspond to vehicles, buses, trucks, bicycles, pedestrians, auto-rickshaws, two-wheelers such as scooters and motorcycles, etc. 

\subsection{Main Contributions}

\begin{enumerate}

\item We present a novel dataset, \rain, corresponding to the dense, heterogeneous, and unstructured traffic in India. \rain~is the first large-scale dataset containing annotated scenes for rare and interesting instances and multi-agent driving behaviors, broadly grouped into:

\begin{enumerate}
    \item Traffic violations---running traffic signals, driving in the wrong lanes, taking wrong turns).
    \item Atypical interactions---cut-ins, yielding, overtaking, overspeeding, zigzagging, lane changing.
    \item Diverse scenarios---intersections, roundabouts, and traffic signals.
\end{enumerate}

\item \rain has more than $2$ million labeled frames and $13$ million annotated bounding boxes for $16$ unique traffic agents, and GPS trajectories for the ego-agent.

\item Every video in \rain is tagged using a diverse range of factors including weather, time of the day, road conditions, and traffic density.

\item We evaluate state-of-the-art methods for object detection and multi-agent behavior prediction on \rain. 

\item We present a novel, fine-grained analysis on the relationship between traffic environments and perception. Specifically we study the effect of 2D object detection in varying traffic density, mixture of agents, area, time of the day, and weather conditions.
\end{enumerate}
\begin{table*}
\centering
\caption{\textbf{Characteristics of Traffic Datasets:} We compare \rain~with state-of-the-art autonomous driving datasets that have been used for trajectory tracking, motion forecasting, semantic segmentation, prediction, and behavior classification. \rain~is the largest (in terms of number of annotated frames) and most diverse in terms of heterogeneity, scenarios, varying behaviors, densities, and rare instances. Darker shades represent a richer collection in that category. Best viewed in color.}
\resizebox{\textwidth}{!}{
\begin{threeparttable}
\begin{tabular}{rcccccccccccc}
\toprule[1.25pt]
  &  &   &  & &  & && && \multicolumn{3}{c}{Rare and Interesting Behaviors$^\ddagger$} \\
 \cmidrule{11-13}
 Datasets & Location & Bad weather & Night & Road type &Het.$^\star$& Size & Density&Lidar&HD Maps& Traffic & Atypical & Diverse \\
   &  & &  & & &  &  & & & Violations & Interactions & Scenarios \\
\midrule

Argoverse~\cite{chang2019argoverse} & USA &\cellcolor{green!25} \cmark&\cellcolor{green!25} \cmark& urban&$10$ &$22$K&Medium&\cellcolor{green!75}\cmark&\cellcolor{green!75}\cmark&\cellcolor{red!25} \xmark&\cellcolor{green!25} \cmark&\cellcolor{red!25} \xmark\\
Lyft Level 5~\cite{lyft} & USA &\cellcolor{red!25} \xmark&\cellcolor{red!25} \xmark &urban&$9$ &$46$K&Low&\cellcolor{green!25}\cmark&\cellcolor{green!25}\cmark&\cellcolor{red!25} \xmark&\cellcolor{red!25} \xmark&\cellcolor{red!25} \xmark\\
 Waymo~\cite{waymo} & USA &\cellcolor{green!25}  \cmark&\cellcolor{green!25} &urban&$4$& $200$K&Medium&\cellcolor{green!75}\cmark&\cellcolor{green!75}\cmark&\cellcolor{red!25} \xmark&\cellcolor{green!25} \cmark&\cellcolor{green!25} \cmark\\
 ApolloScape~\cite{apolloscape}& China &\cellcolor{red!25} \xmark &\cellcolor{green!25} \cmark &\cellcolor{cyan!25} urban, rural &$5$ &$144$K&\cellcolor{cyan!25} High&\cellcolor{green!25}\cmark&\cellcolor{green!25}\cmark&\cellcolor{red!25} \xmark&\cellcolor{red!25} \xmark&\cellcolor{red!25} \xmark\\
 nuScenes~\cite{nuscenes2019}& USA/Sg.&\cellcolor{green!25} \cmark&\cellcolor{green!25} \cmark & urban&$13$ &$40$K&Low&\cellcolor{green!75} \cmark&\cellcolor{green!75} \cmark&\cellcolor{red!25} \xmark&\cellcolor{green!25} \cmark&\cellcolor{green!25} \cmark\\
 INTERACTION~\cite{zhan2019interaction}& International&\cellcolor{red!25} \xmark&\cellcolor{red!25} \xmark & \cellcolor{cyan!25} urban&$1$ &$-$&Medium&\cellcolor{green!25} \cmark&\cellcolor{green!25} \cmark&\cellcolor{red!25} \xmark&\cellcolor{red!25} \xmark&\cellcolor{red!25} \xmark\\


CityScapes~\cite{cordts2016cityscapes} &Europe&\cellcolor{red!25} \xmark &\cellcolor{red!25} \xmark &urban& $10$ &$25$K&Low&\cellcolor{red!25} \xmark&\cellcolor{red!25} \xmark&\cellcolor{red!25} \xmark&\cellcolor{red!25} \xmark&\cellcolor{red!25} \xmark\\
IDD~\cite{varma2019idd} &India&\cellcolor{red!25} \xmark&\cellcolor{red!25} \xmark &\cellcolor{cyan!75} urban, rural&$12$ &$10$K& \cellcolor{cyan!75} High&\cellcolor{red!25} \xmark&\cellcolor{red!25} \xmark&\cellcolor{red!25} \xmark&\cellcolor{red!25} \xmark&\cellcolor{red!25} \xmark\\


HDD~\cite{hdd} & USA &\cellcolor{red!25} \xmark&\cellcolor{red!25} \xmark & urban& $-$&$275$K&Medium&\cellcolor{green!25} \cmark&\cellcolor{red!25}\xmark&\cellcolor{red!25} \xmark&\cellcolor{green!25}\cmark&\cellcolor{green!25}\cmark\\
Brain4cars~\cite{jain2016brain4cars} & USA &\cellcolor{red!25} \xmark&\cellcolor{red!25} \xmark & urban&$-$ &$2000$K&Low&\cellcolor{red!25} \xmark&\cellcolor{green!25} \cmark&\cellcolor{red!25} \xmark&\cellcolor{red!25} \xmark&\cellcolor{red!25} \xmark\\
D2-City~\cite{d2city} & China &\cellcolor{green!25} \cmark&\cellcolor{red!25} \xmark & urban&$12$ &$700$K&Medium&\cellcolor{red!25} \xmark&\cellcolor{red!25} \xmark&\cellcolor{red!25} \xmark&\cellcolor{red!25} \xmark&\cellcolor{green!25}\cmark\\
TRAF~\cite{chandra2019traphic} &India&\cellcolor{red!25} \xmark&\cellcolor{green!25} \cmark &\cellcolor{cyan!75} urban, rural&$8$ &$72$K& \cellcolor{cyan!75} High&\cellcolor{red!25} \xmark&\cellcolor{red!25} \xmark&\cellcolor{red!25} \xmark&\cellcolor{red!25} \xmark&\cellcolor{red!25} \xmark\\
BDD~\cite{chen2018bdd100k} &USA&\cellcolor{green!25} \cmark&\cellcolor{green!25} \cmark &urban&$8$ &$\bm{3000}$K&Low&\cellcolor{red!25} \xmark&\cellcolor{red!25} \xmark&\cellcolor{red!25} \xmark&\cellcolor{red!25} \xmark&\cellcolor{green!25} \cmark\\
\midrule

 \textbf{\rain} & \textbf{India} &\cellcolor{green!75} \cmark&\cellcolor{green!75} \cmark &\cellcolor{cyan!75} \textbf{urban, rural}$^\dagger$&$\bm{16}^{\dagger\dagger}$ &$2027$K&\cellcolor{cyan!75} \textbf{High}$^\mathsection$ &\cellcolor{red!25} \xmark&\cellcolor{red!25} \xmark&\cellcolor{green!75} \cmark&\cellcolor{green!75} \cmark&\cellcolor{green!75} \cmark\\

\bottomrule[1.25pt]
\end{tabular}
\begin{tablenotes}\footnotesize
\item[$\ddagger$] Rare instances can be broadly grouped into $(i)$ traffic violations, $(ii)$ atypical interactions, and $(iii)$ difficult scenarios.
\item[$\dagger$] Includes roads without lane markings. Roads in other datasets with rural roads may contain lane markings.
\item[$\star$] Heterogeneity. We indicate the classes corresponding to moving traffic agents only, excluding static objects such as poles, traffic lights, etc. 
\item[$\mathsection$] Up to $40$ agents per frame.
\item[$\dagger\dagger$] Up to $9$ unique agents per frame.

\end{tablenotes}
\end{threeparttable}
}
\label{tab: related_work}
\vspace{-10pt}
\end{table*}

\subsection{Applications and Benefits}

\begin{itemize}
    \item \textbf{Towards Risk-Aware Planning and Control:} Our multi-agent behavior prediction benchmark can aid the development of risk-aware motion planners by predicting the behaviors of surrounding agents. Motion planners can compute controls that guarantee safety around aggressive drivers who are prone to overtaking and overspeeding.
    
    \item \textbf{Towards Robust Perception:} We observe that these models fail in challenging Indian traffic scenarios, compared to their performance on existing datasets captured in the US, Europe, and other developed nations. As a result, \rain can be a useful benchmark for research in perception in unstructured traffic environments and developing nations.
    \item \textbf{Towards Fine-grained Traffic Analysis:} Our novel analysis studying the relationship between traffic patterns and 2D object detection can lead to more informed research in perception for autonomous driving.
\end{itemize}

\section{Comparison with Existing Datasets}
\label{sec: related_work}
\subsection{Tracking and Trajectory Prediction Datasets}
\label{subsec: trajectory_prediction_related}

Datasets such as the Argoverse~\cite{chang2019argoverse}, Lyft Level 5~\cite{lyft}, Waymo Open Dataset~\cite{waymo}, ApolloScape~\cite{apolloscape}, nuScenes dataset~\cite{nuscenes2019} are used for trajectory forecasting~\cite{chandra2019traphic, chandra2019robusttp, chandra2020forecasting, zhao2020tnt, chai2019multipath} and tracking~\cite{chandra2019roadtrack}. Several of these datasets use mining procedure~\cite{waymo, chang2019argoverse} that heuristically searches the dataset for rare and interesting scenarios. The resulting collection of such scenarios and behaviors, however, is only a fraction of the entire dataset. \rain, by comparison, exclusively contains such scenarios due to the inherent nature of the unstructured traffic in India.

\rain~has many additional characteristics with respect to these datasets. For instance, \rain's $2.02$ million annotated frames are more than $10\times$ the current highest number of annotated frames with respect to other dataset with high density traffic (ApolloScape). Furthermore, \rain~consists of $16$ different traffic-agents that include only on-road moving entities (and not static obstacles). This is by far, the most diverse in terms of class labels. In comparison, Argoverse and nuScenes both contain $10$ and $13$ traffic-agents, respectively. \rain~is the first motion forecasting and behavior prediction dataset with traffic patterns from rural and urban areas that consist of unmarked roads and high-density traffic. In contrast, traffic scenarios in Argoverse, Waymo, Lyft, and nuScenes have been captured on sparse to medium density traffic with well-marked structured roads in urban areas. 


\subsection{Semantic Segmentation Datasets}
\label{subsec: semantic_segmentation_related}

CityScapes~\cite{cordts2016cityscapes} is widely used for several tasks, primarily semantic segmentation. It is based on urban traffic data collected from European cities with structured roads and low traffic density. In contrast, the Indian Driving Dataset (IDD)~\cite{varma2019idd} is collected in India with both urban and rural areas with high-density traffic. A common aspect of both these  datasets (CityScapes and IDD), however, is the relatively low annotated frame count ($25$K and $10$K, respectively). This is probably due to the effort involved with annotating every pixel in each image. 
IDD also contains high-density traffic scenarios in rural areas, similar to \rain. However, our dataset has $200\times$ the number of annotated frames and $1.6\times$ the number of traffic-agent classes. Similar to TRAF, the IDD does not contain the behavior data that is provided by \rain.


\subsection{Behavior Prediction}
\label{subsec: action_prediction_related}

Behavior prediction corresponds to the task of predicting turns (right, U-turn, or left), acceleration, merging, and braking in addition to driver-intrinsic behaviors such as over-speeding, overtaking, cut-ins, yielding, and rule-breaking. The two most prominent datasets for action prediction include the Honda Driving Dataset (HDD)~\cite{hdd} and the BDD dataset~\cite{chen2018bdd100k}. Some of the major distinctions between \rain~and the HDD in terms of size (approximately $10\times$), the availability of scenes with night driving and rainy weather, and the inclusion of unstructured environments in low-density traffic. The BDD dataset~\cite{chen2018bdd100k} contains more annotated samples than \rain, however, the BDD dataset contains $100$K videos while \rain~contains $1$K videos. So the number of annotated samples per video is $66\times$ higher for \rain. The annotations in prior datasets are limited to actions and do not contain the rare and interesting behaviors contained in \rain.

\section{\rain~dataset}
\label{sec: dataset}

\begin{figure*}[t]
\centering   
\begin{adjustbox}{minipage=\textwidth,scale=1}
   \begin{subfigure}[h]{0.492\textwidth}
    \includegraphics[width=\textwidth, height = 2.32cm]{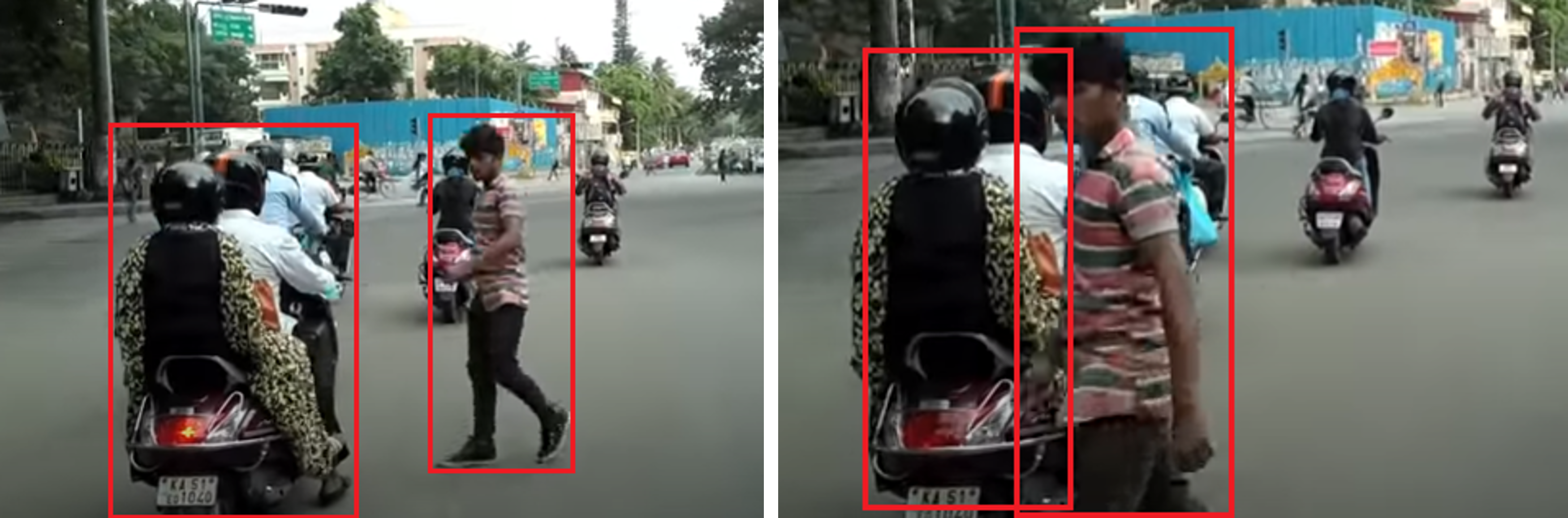}
    \caption{Cut-ins/Jaywalking.}
    \label{fig: a}
  \end{subfigure}
 \begin{subfigure}[h]{0.492\textwidth}
    \includegraphics[width=\textwidth]{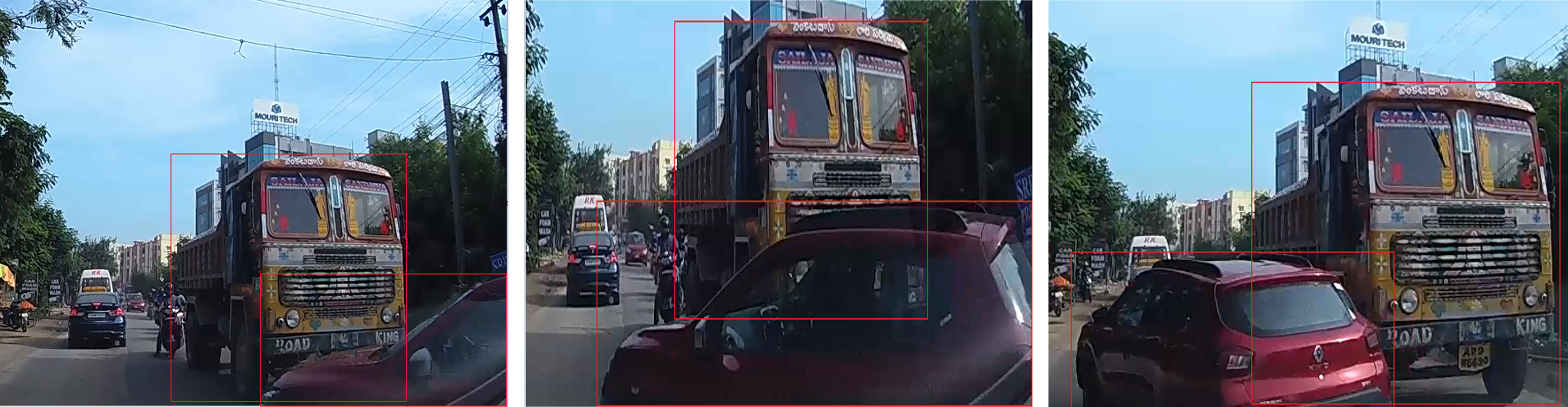}
    \caption{Yielding/Cut-ins.}
    \label{fig: b}
  \end{subfigure}
   \begin{subfigure}[h]{0.492\textwidth}
    \includegraphics[width=\textwidth, height=2.32cm]{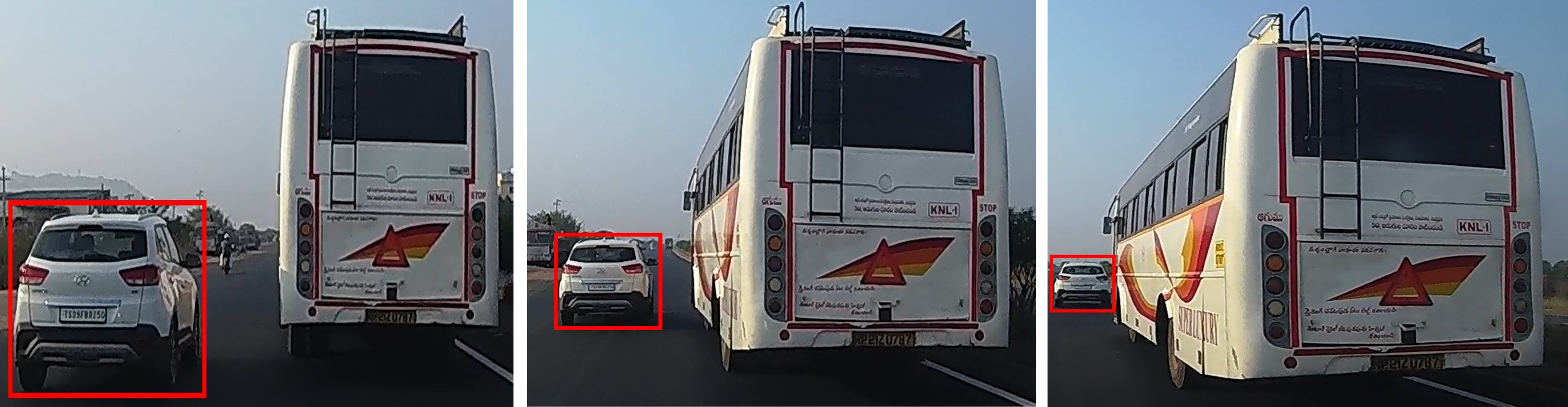}
    \caption{Overtaking/Overspeeding.}
    \label{fig: c}
  \end{subfigure}
 \begin{subfigure}[h]{0.492\textwidth}
    \includegraphics[width=\textwidth, height = 2.32cm]{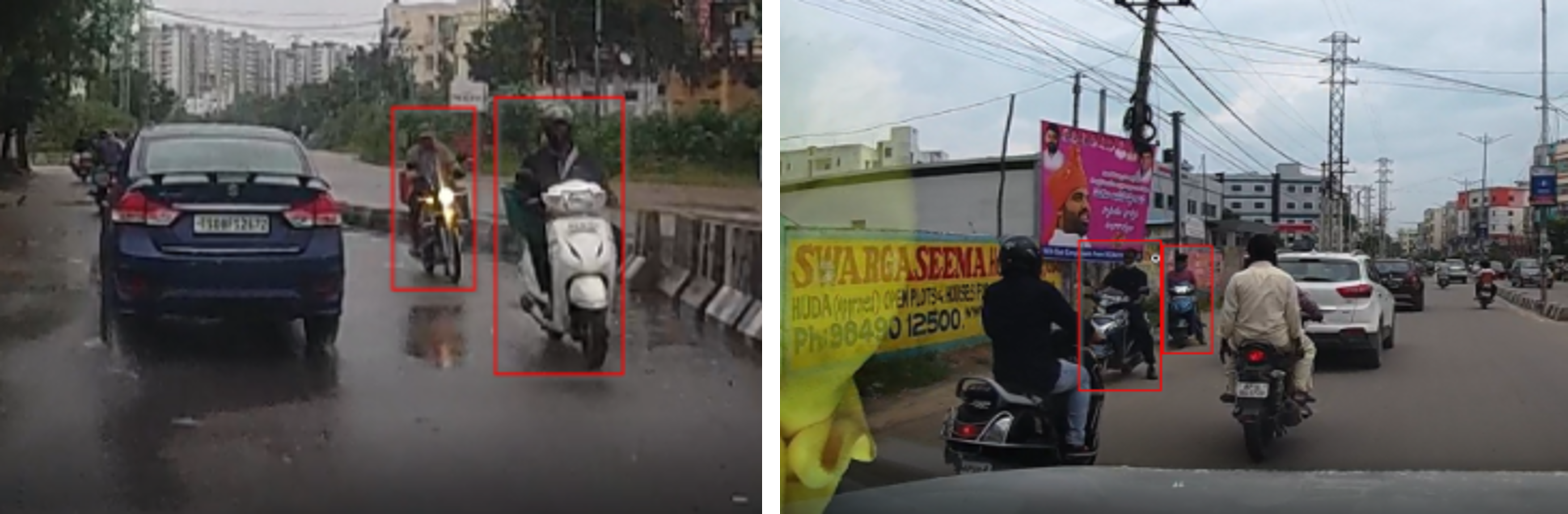}
    \caption{Driving in wrong lane.}
    \label{fig: d}
  \end{subfigure}
 \begin{subfigure}[h]{0.492\textwidth}
    \includegraphics[width=\textwidth,height = 2.32cm]{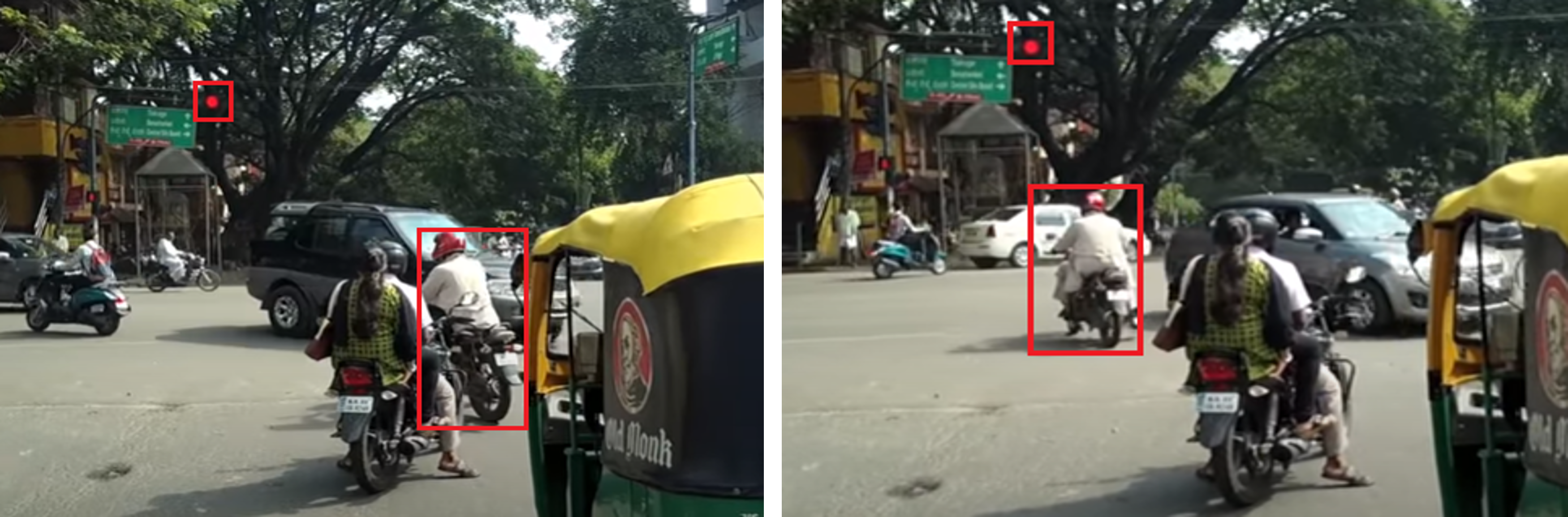}
    \caption{Running red traffic lights.}
    \label{fig: e}
  \end{subfigure}
   \begin{subfigure}[h]{0.492\textwidth}
    \includegraphics[width=\textwidth, height=2.32cm]{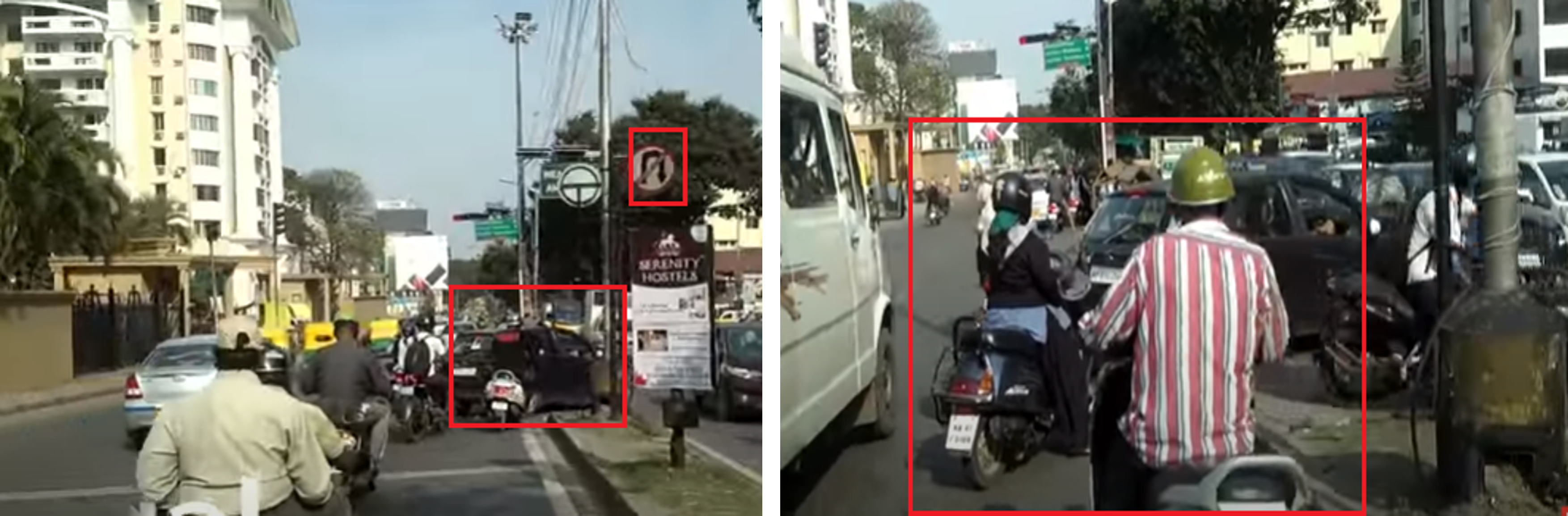}
    \caption{Ignoring lane signs/wrong lane driving.}
    \label{fig: f}
  \end{subfigure}
   \begin{subfigure}[h]{0.245\textwidth}
    \includegraphics[width=\textwidth]{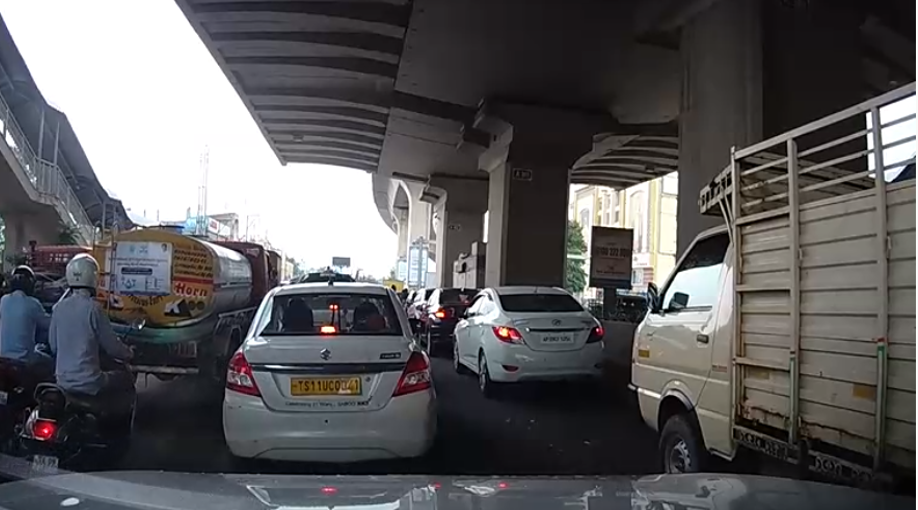}
    \caption{High density.}
    \label{fig: g}
  \end{subfigure}
 \begin{subfigure}[h]{0.245\textwidth}
    \includegraphics[width=\textwidth]{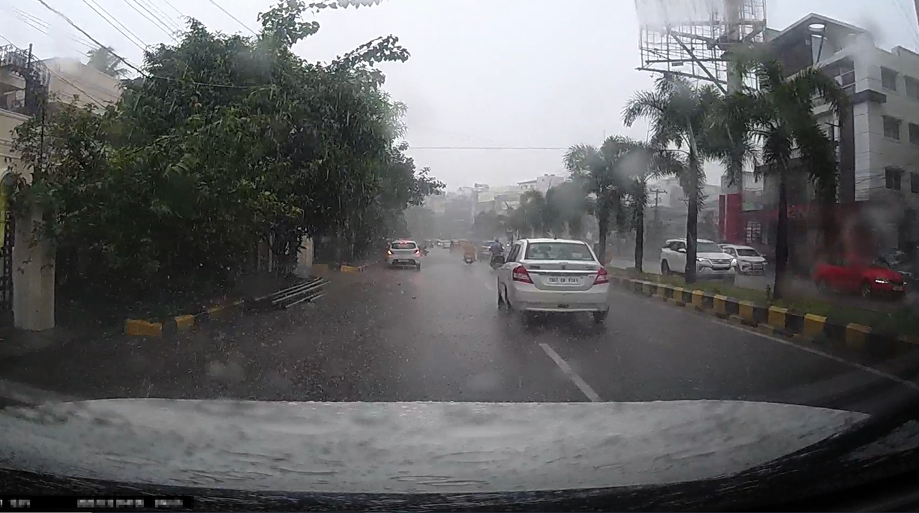}
    \caption{Rainy weather.}
    \label{fig: h}
  \end{subfigure}
   \begin{subfigure}[h]{0.245\textwidth}
    \includegraphics[width=\textwidth, height=2.32cm]{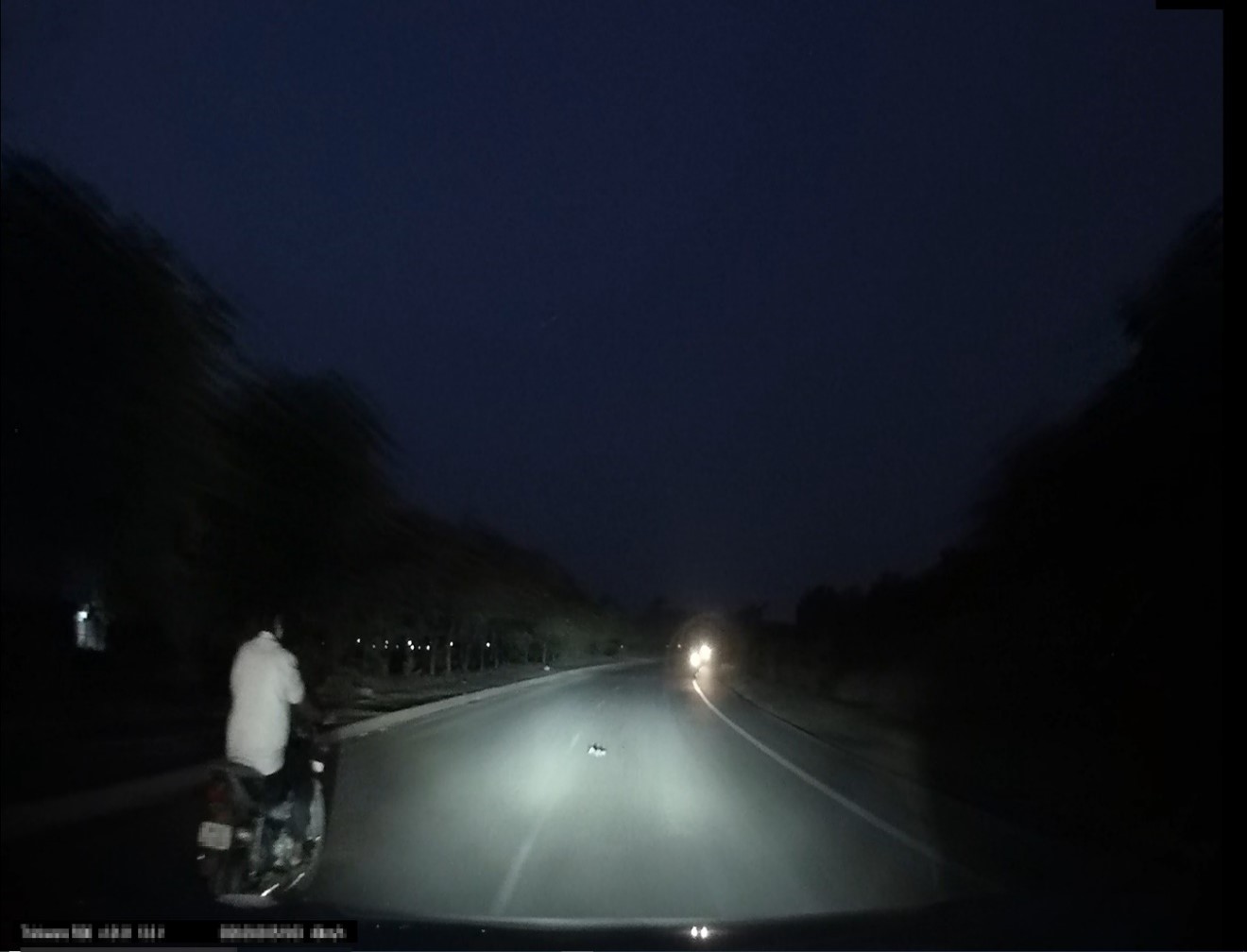}
    \caption{Night time.}
    \label{fig: i}
  \end{subfigure}
 \begin{subfigure}[h]{0.245\textwidth}
    \includegraphics[width=\textwidth]{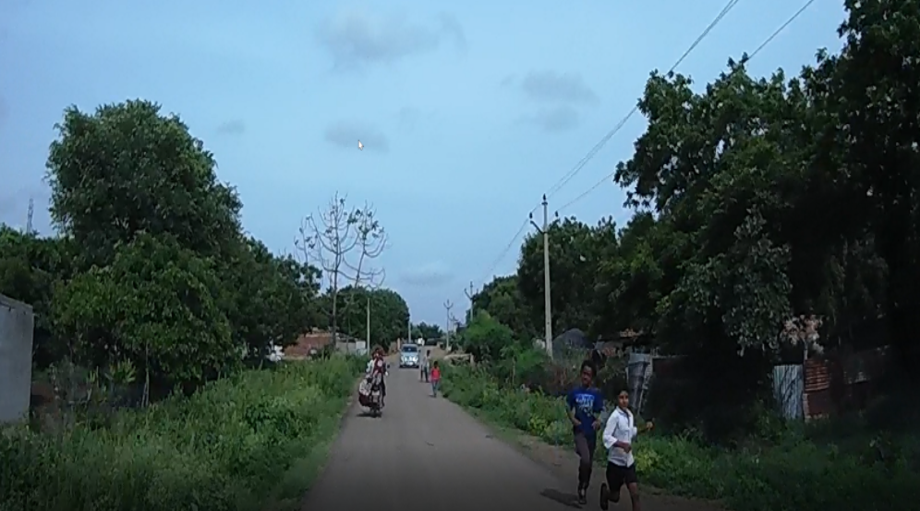}
    \caption{Rural areas.}
    \label{fig: j}
  \end{subfigure}
\end{adjustbox}

\caption{\textbf{Annotations for rare instances:} One of the unique aspects of \rain~is the availability of explicit labels for rare and interesting instances including atypical interactions, traffic violations, and diverse scenarios. These annotations can be used to benchmark new methods for object detection and multi-agent behavior prediction.}
  \label{fig: rulebreak}
  \vspace{-10pt}
\end{figure*}

Our dataset is visually shown in Figure~\ref{fig: infographic}. Below, we present some details of the data collection process and discuss some of the salient features and characteristics of \rain.

\subsection{Dataset Collection}

The data was collected in and around the city of Hyderabad, India within a radius of $42$ to $62$ miles. Several outskirts were chosen to cover rural and unstructured roads. Our hardware capture setup consists of two wide-angle Thinkware F800 dashcams mounted on an MG Hector and Maruti Ciaz. The camera sensor has  $2.3$ megapixel resolution with a $140\degree$ field of view. The video is captured in full high definition with a resolution of $1920\times1080$ pixels at a frame rate of $30$ frames per second. The dashcam is embedded with an accurate positioning system that stores the GPS coordinates, which were processed into the world frame coordinates. The sensor synchronizes between the camera and the GPS. Recordings from the dashcam are streamed continuously and are clipped into $1$ minute video segments.

\subsection{Dataset organization}
The dataset is organized as $1250$ one-minute video clips. Each clip contains static and dynamic XML files. Each  static file summarizes the meta-data of the entire video clip including the behaviors, road type, scene structure etc. Each dynamic file describes frame-level information such as bounding boxes, GPS coordinates, and agent behaviors. Our dataset can be searched using helpful filters that sort the data according to the road type, traffic density, area, weather, and behaviors. We also provide many scripts to easily load the data after downloading.

\subsection{Annotations}

We provide the following annotations in our dataset: $(i)$ bounding boxes for every agent, $(ii)$ agent class IDs, $(iii)$ GPS trajectories for the ego-vehicle, $(iv)$ environment conditions including weather, time of the day, traffic density, and heterogeneity, $(v)$ road conditions with urban, rural, lane markings, $(vi)$ road network including intersections, roundabouts, traffic signal, $(vii)$ actions corresponding to left/right turns, U-turns, accelerate, brake, $(viii)$ rare and interesting behaviors (See Section~\ref{subsec: rare}), and $(ix)$ the camera intrinsic matrix for depth estimation to generate trajectories of the surrounding vehicles. This set of annotations is the most diverse and extensive compared prior datasets.

A diverse and rich taxonomy of agent categories is necessary to ensure that autonomous driving systems can detect different types of agents in any given scenario. Towards that goal, datasets for autonomous driving are designed or captured to achieve two goals: $(a)$ capture as many different types of agent categories as possible; $(b)$ capture as many instances of each category as possible. In both these aspects, \rain~outperforms all prior datasets. We annotate $16$ types of moving traffic entities, not including static obstacles listed in Figure~\ref{fig: infographic} along with their distribution. Note specifically that  the percentages of pedestrians, motorbikes, and bicycles are higher than the percentage of passenger vehicles. This is particularly useful as the former categories are known as ``vulnerable road users'' (VRUs)~\cite{vru}, and it is important for autonomous driving systems to be able to detect them--necessitating many instances of these VRUs in any dataset.






\subsection{Rare and Interesting Behaviors}
\label{subsec: rare}

We provide a total of $17$ different types of rich collection of rare and interesting cases that are unique to our dataset. They can be summarized in terms of the following groups:

\subsubsection{Atypical Interactions}
Atypical interactions correspond to pairwise interactions among traffic agents that are not often observed in regular traffic scenarios. Some examples of atypical interactions include yielding to, and cutting across, pedestrians, zigzagging through traffic, pedestrian jaywalking, overtaking, sudden lane changing, and overspeeding. We describe these in more detail below:

\begin{itemize}
\item \textit{Overtaking (OT)}: When an agent overtakes another agent with sudden or aggressive movement.
\item \textit{Overspeeding (OS)}: If the vehicle over-speeds (based on speed limits) due to any reason.
\item \textit{Yield (Y)}: A pedestrian, bicycle, or any slow-moving agent trying to cross the road in front of another agent. If the latter slows down or stops, letting them cross the road then such behavior is labeled as yield.
\item \textit{Cutting (C)}: When pedestrians, bicycles, or any slow-moving agents trying to cross the road is interrupted by another agent. Yielding and cutting can also be re-labeled as instances of jaywalking. In a majority of these cases, one of the agents involved is a pedestrian crossing the road in the middle of traffic.

\item \textit{Lane change w. lane markings (LC(m))}: Agents aggressively change lanes on roads with clear lane markings.
\item \textit{Lane change w/o. lane markings (LC)}: Agents aggressively change lanes on roads without lane markings.

The above two annotations can be used to identify videos in the dataset that contain roads without lane markings for relevant applications.

\item \textit{Zigzagging (ZM)}: If any of the agent of interest undergoes a zigzag movement in the traffic, the agent behavior is classified as zigzagging.

\end{itemize}

\subsubsection{Traffic Violations}

In addition to the above driving behaviors, we also annotate traffic agents breaking traffic rules. These are particularly unique since rule breaking scenarios are rare.
\begin{itemize}
\item \textit{Running a traffic light (RB TL)}: Passing through an intersection even though the traffic signal is red. 
\item \textit{Wrong Lane (RB WL)}: A road may not be divided for inbound and outbound traffic by a physical barrier, making it possible for the motorists to use the inbound lane for the outbound traffic and vice versa. This behavior identifies all such cases.
\item \textit{Wrong Turn (RB WT)}: When an agent makes an illegal turn (including U-turns).
\end{itemize}

\subsubsection{Diverse Scenarios}

Finally, we provide annotations for challenging scenarios that include intersections, roundabouts, traffic signals, executing left turns, right turns, and U-turns.

\subsection{Dataset statistics}
\label{subsec: dataset_stats}
\begin{figure}[t]
\centering
   \begin{subfigure}[h]{\columnwidth}
    \includegraphics[width=\textwidth]{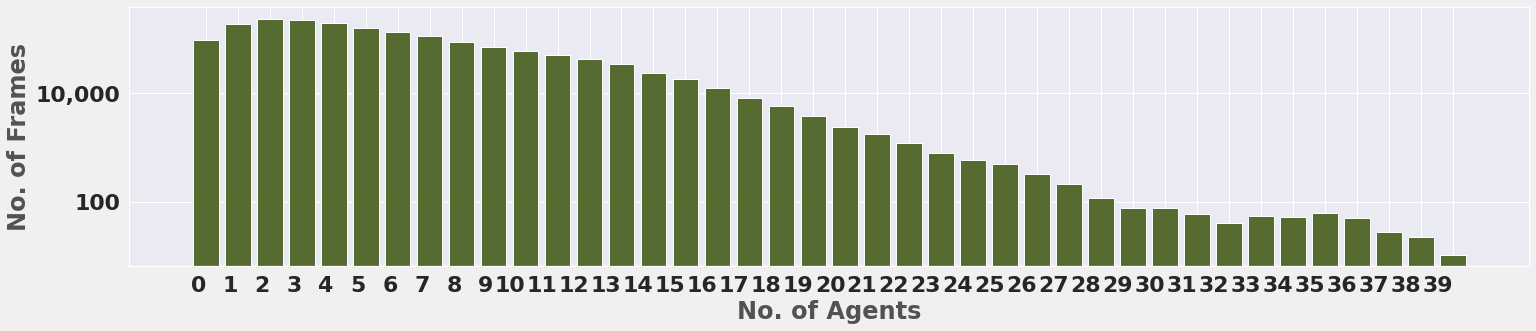}
    \caption{\textbf{High traffic density:} \rain~has up to $40$ agents per frame. }
    \label{fig: total_agents_per_frame}
        \vspace{3pt}
  \end{subfigure}
  %
   \begin{subfigure}[h]{.45\columnwidth}
 \centering
    \includegraphics[width=\textwidth]{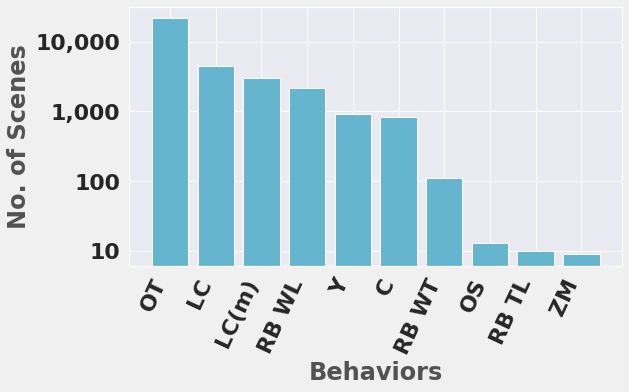}
    \caption{Number of scenes in which behaviors occur}
    \label{fig: number_of_scenes_behaviors}
  \end{subfigure}
  \begin{subfigure}[h]{.45\columnwidth}
  \centering
    \includegraphics[width=\textwidth, height=2.5cm]{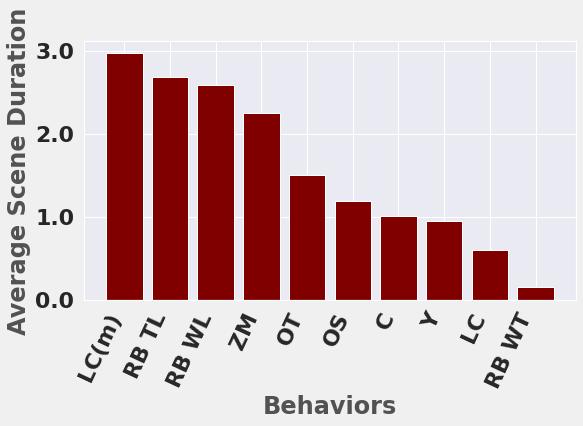}
    \caption{Average scene duration of behaviors (in seconds)}
    \label{fig: scene_duration}
  \end{subfigure}
 \begin{subfigure}[h]{.45\columnwidth}
    \includegraphics[width=\textwidth, height=2.3cm]{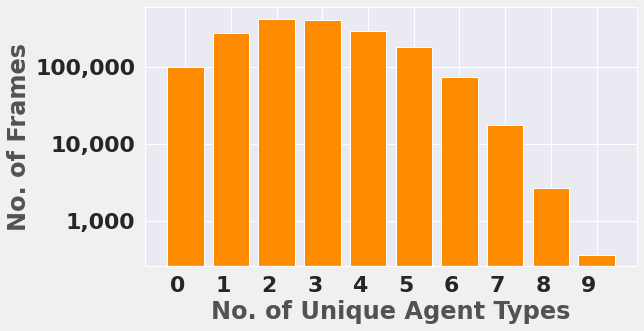}
    \caption{\textbf{High heterogeneity:} Up to $9$ unique agents in a single frame.}
    \label{fig: unique_agents_per_frame}
  \end{subfigure}
 \begin{subfigure}[h]{.45\columnwidth}
    \includegraphics[width=\textwidth, height=2.3cm]{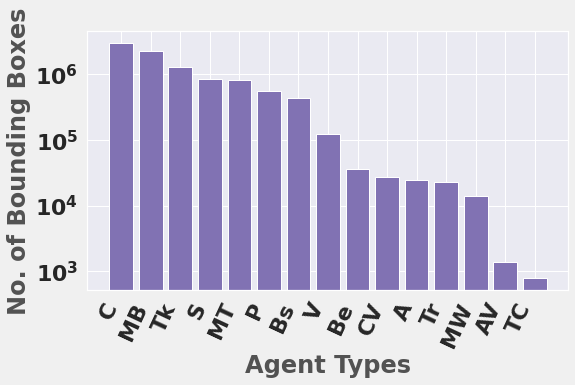}
    \caption{\textbf{Rich features:} Up to $13$ million boxes.}
    \label{fig: bounding_box}
  \end{subfigure}
  
\caption{We highlight the high traffic density, heterogeneity, and the richness of behavior information in \rain. Abbreviations correspond to various behavior categories and are explained in Section~\ref{subsec: rare}.}
  \label{fig: stats}
  \vspace{-10pt}
\end{figure}

We analyze the dataset statistics and distribution of agents and their behaviors in terms of total count, uniqueness, and duration (in seconds). Figures~\ref{fig: total_agents_per_frame} and~\ref{fig: unique_agents_per_frame} show that \rain is very dense and highly heterogeneous, respectively; the total number of agents in a single frame can reach up to $40$ and up to $9$ unique agents can exist in a single frame. Figure~\ref{fig: number_of_scenes_behaviors} represents the distribution of behaviors across videos and Figure~\ref{fig: scene_duration} shows the distribution of each behavior's average duration. In particular, we note that the average duration can reach up to $3$ seconds which, at $30$ frames per second, corresponds to approximately $90$ frames that contain visual, contextual, and semantic information that can inform behavior prediction algorithms for more accurate perception and prediction.

\section{Experiments and Analysis}
We provide the pre-trained models for object detection and behavior prediction at \site.

\begin{table*}[t]
\centering
\caption{\textbf{Effect of meta features on object detection:} We analyze how meta features such as traffic density, type of agents, location, time of the day, and weather play a role in 2D object detection using the DETR, Deformable DETR, YOLOv3 and CenterNet object detectors. \textbf{Bold} indicates the type of meta feature that is the most effective for object detection.}
\resizebox{\textwidth}{!}{
\begin{tabular}{lccccccccccc}
\toprule[1.25pt]
\multicolumn{12}{c}{DETR and Deformable DETR (in parentheses)}\\
\midrule
 &  \multicolumn{3}{c}{Density}&  \multicolumn{2}{c}{Agents} &  \multicolumn{2}{c}{Environment} &  \multicolumn{2}{c}{Time} &  \multicolumn{2}{c}{Weather} \\
\cmidrule{2-12}
&  \small{Low} &  \small{\textbf{Medium}} &  \small{High} &  \small{\textbf{Mixed}} &  \small{Uniform} &  \small{\textbf{Urban}} &  \small{Rural} &  \small{\textbf{Day}} &  \small{Night} &  \small{\textbf{Normal}} &  \small{Rainy} \\
\cmidrule{2-12}
mAP               & $19.00$ ($22.70$)& $27.00$ ($38.30$)& $19.30$ ($28.10$)& $27.00$ ($38.30$)& $14.80$ ($31.30$)& $27.00$ ($38.30$)& $14.20$ ($25.70$)& $27.00$ ($38.30$)& $12.00$ ($20.60$)& $27.00$ ($38.30$)& $12.00$ ($20.90$)\\
mAP$_{50}$        & $33.33$ ($36.80$)& $48.40$ ($61.80$)& $32.40$ ($41.40$)& $48.40$ ($61.80$)& $31.80$ ($44.30$)& $48.40$ ($61.80$)& $23.40$ ($34.90$)& $48.40$ ($61.80$)& $22.70$ ($36.10$)& $48.40$ ($61.80$)& $21.90$ ($32.70$)\\
mAP$_{75}$        & $21.50$ ($22.10$)& $28.10$ ($41.50$)& $20.40$ ($31.30$)& $28.10$ ($41.50$)& $11.70$ ($37.00$)& $21.80$ ($41.50$)& $16.30$ ($28.40$)& $28.10$ ($41.50$)& $12.20$ ($20.50$)& $28.10$ ($41.50$)& $12.60$ ($22.90$)\\
mAP$_{\textrm{S}}$  & $ 2.60$ ($7.10$)& $ 1.20$ ($12.10$)& $ 0.20$ ($2.50$)& $ 1.20$ ($12.10$)& $ 0.30$ ($12.80$)& $ 1.20$ ($12.10$)& $ 2.00$ ($10.30$)& $ 1.20$ ($12.10$)& $ 0.10$ ($0.30$) & $1.20$ ($12.10$)& $1.80$ ($9.50$)\\
mAP$_{\textrm{M}}$  & $ 7.40$ ($25.20$)& $ 8.30$ ($22.50$)& $10.50$ ($16.90$)& $ 8.30$ ($22.50$)& $ 7.20$ ($34.30$)& $ 8.30$ ($22.50$)& $11.70$ ($28.10$) & $ 8.30$ ($22.50$)&$ 3.30$ ($12.50$)&$8.30$ ($22.50$) & $6.20$ ($19.90$)\\
mAP$_{\textrm{L}}$  & $25.60$ ($24.90$)& $45.90$ ($54.10$)& $24.70$ ($35.60$)& $45.90$ ($54.10$)& $40.30$ ($57.80$)& $45.90$ ($54.10$)& $26.30$ ($35.60$)& $45.90$ ($54.10$)& $16.70$ ($27.80$)& $45.90$ ($54.10$)& $15.10$ ($23.80$)\\
\bottomrule[1.25pt]
\toprule[1.25pt]
\multicolumn{12}{c}{YOLOv3 and CenterNet (in parentheses)}\\
\midrule
 &  \multicolumn{3}{c}{Density}&  \multicolumn{2}{c}{Agents} &  \multicolumn{2}{c}{Environment} &  \multicolumn{2}{c}{Time} &  \multicolumn{2}{c}{Weather} \\
\cmidrule{2-12}
&  \small{Low} &  \small{\textbf{Medium}} &  \small{High} &  \small{\textbf{Mixed}} &  \small{Uniform} &  \small{\textbf{Urban}} &  \small{Rural} &  \small{\textbf{Day}} &  \small{Night} &  \small{\textbf{Normal}} &  \small{Rainy} \\
\cmidrule{2-12}
mAP               & $19.20$ ($22.90$)& $30.40$ ($32.90$)& $21.10$ ($23.30$)& $30.40$ ($32.90$)& $19.10$ ($30.20$)& $30.40$ ($32.90$)& $13.80$ ($13.60$)& $30.40$ ($32.90$)& $13.30$ ($15.90$)& $30.40$ ($32.90$)& $13.40$ ($14.00$)\\
mAP$_{50}$        & $36.90$ ($34.80$)& $52.50$ ($55.40$)& $36.30$ ($32.50$)& $52.50$ ($55.40$)& $35.10$ ($43.40$)& $52.50$ ($55.40$)& $22.00$ ($22.70$)& $52.50$ ($55.40$)& $25.00$ ($25.70$)& $52.50$ ($55.40$)& $25.00$ ($22.50$)\\
mAP$_{75}$        & $16.10$ ($28.10$)& $32.30$ ($33.40$)& $23.20$ ($26.70$)& $32.30$ ($33.40$)& $19.70$ ($37.30$)& $32.30$ ($33.40$)& $15.70$ ($13.20$)& $32.30$ ($33.40$)& $13.40$ ($27.00$)& $32.30$ ($33.40$)& $13.60$ ($15.50$)\\
mAP$_{\textrm{S}}$& $ 2.70$ ($ 8.40$)& $ 2.40$ ($13.10$)& $ 0.60$ ($ 2.90$)& $ 2.40$ ($13.10$)&       $ 7.90$ ($19.30$)& $ 2.40$ ($13.10$)& $ 5.20$ ($ 5.40$)& $ 2.40$ ($13.10$)& $ 0.00$ ($ 0.90$) & $2.40$ ($13.10$)& $1.30$ ($10.90$)\\
mAP$_{\textrm{M}}$& $14.10$ ($26.20$)& $13.10$ ($30.50$)& $11.70$ ($17.60$)& $ 13.10$ ($30.50$)& $19.10$ ($38.80$)& $13.10$ ($30.50$)& $22.50$ ($25.80$)& $13.10$ ($30.50$)& $ 7.50$ ($11.60$)& $13.10$ ($30.50$) & $11.60$ ($17.40$)\\
mAP$_{\textrm{L}}$& $23.70$ ($29.50$)& $48.70$ ($44.60$)& $27.30$ ($27.90$)& $48.70$ ($44.60$)& $38.90$ ($40.00$)& $48.70$ ($44.60$)& $21.20$ ($21.40$)& $48.70$ ($44.60$)& $18.50$ ($21.70$)& $48.70$ ($44.60$)& $16.40$ ($14.30$)\\
\bottomrule[1.25pt]
\end{tabular}
}
\label{tab: meta_od}
\end{table*}

\subsection{Analyzing Object Detection in Unstructured Scenarios}
\label{subsec: OD}
\begin{table}[t]
    \centering
    \caption{\textbf{Training Details for Object Detection} (BS: Batch size, Mom: Momentum, WD: Weight decay, MGN: Max Gradient Norm)}
    \resizebox{\columnwidth}{!}{
    \begin{tabular}{rccccccc}
\toprule[1.25pt]
 Method & Backbone & BS & Opt. & LR & Mom. & WD ($L_{2}$) & MGN\\
\midrule
DETR~\cite{od-model} & ResNet-50 & $2$ & AdamW & $1\mathrm{e}{-4}$ & $-$ & $1\mathrm{e}{-4}$ & $0.1$ \\
Def. DETR~\cite{zhu2020deformable} & ResNet-50 & $2$ & AdamW & $2\mathrm{e}{-4}$ & $-$ & $1\mathrm{e}{-4}$ & $0.1$ \\
YOLOv3~\cite{redmon2018yolov3} & Darknet-53 & $8$ & SGD & $1\mathrm{e}{-3}$ & $0.9$ & $5\mathrm{e}{-4}$ & $35$ \\
CenterNet~\cite{centernet} & ResNet-18 & $16$ & SGD & $1\mathrm{e}{-3}$ & $0.9$ & $5\mathrm{e}{-4}$ & $35$ \\
\bottomrule[1.25pt]
\end{tabular}
}
\label{tab: od_hyperparams}
\vspace{-10pt}
\end{table}

\begin{table}[t]
\centering
\caption{\textbf{Object detection on Waymo and KITTI:} We report the standard mAP for many widely used methods on autonomous driving datasets.}
\resizebox{\columnwidth}{!}{
\begin{tabular}{rccccc}
\toprule[1.25pt]
 & DETR~\cite{od-model} & CenterNet & YOLO v3 & Def. DETR & Swin-T\\
\cmidrule{2-6}
KITTI~\cite{kitti} & $23.00$ & $80.40$ & $81.60$ & $42.20$ & $-$ \\
Waymo~\cite{waymo} & $65.31$ & $64.83$& $56.93$ & $65.31$  & $37.20$ \\
\midrule
\textbf{METEOR} & $\bm{8.30}$ & $\bm{12.10}$ & $\bm{14.30}$ & $\bm{15.80}$ & $\bm{32.60}$ \\
\bottomrule[1.25pt]
\end{tabular}
}
\label{tab: mAP_table}
\vspace{-10pt}

\end{table}

\begin{table}[t]
\centering
\caption{\textbf{Swin-T on Waymo and METEOR:} We present a more detailed analysis of Swin-T, one of the state-of-the-art object detection approaches, on Waymo and METEOR.}  
\resizebox{\columnwidth}{!}{
\begin{tabular}{rcccccc}
\toprule[1.25pt]
& mAP & mAP$_{50}$ & mAP$_{75}$ & mAP$_{\textrm{S}}$ & mAP$_{\textrm{M}}$ & mAP$_{\textrm{L}}$\\
\cmidrule{2-7}
Waymo~\cite{waymo} & $37.20$ & $70.60$ & $52.00$ & $17.20$ & $41.80$ & $67.20$\\
\midrule
\textbf{\rain} & $\bm{32.60}$ & $\bm{46.90}$ & $\bm{36.20}$ & $\bm{20.50}$ & $\bm{35.40}$ & $\bm{54.70}$\\
\bottomrule[1.25pt]
\end{tabular}
}
\label{tab: swint_table}
\vspace{-10pt}

\end{table}

Existing datasets have helped develop sophisticated and robust 2D detection methods. We use the MMDetection~\cite{mmdetection} toolbox to train the following 2D object detection models---DETR~\cite{od-model}, Deformable DETR~\cite{zhu2020deformable} (with iterative bounding box refinement), YOLOv3~\cite{redmon2018yolov3} (with scale $608$), CenterNet~\cite{centernet} (with normal convolutions), and Swin-T~\cite{liu2021swin}. The models are pre-trained on the COCO dataset~\cite{coco} and fine-tuned on \rain. We provide the training details in Table \ref{tab: od_hyperparams} and report results using the standard mAP, mAP$_{50}$, mAP$_{75}$, mAP$_{\textrm{S}}$, mAP$_{\textrm{M}}$, and mAP$_{\textrm{L}}$. We refer the reader to~\cite{od_metrics} for a primer on these metrics.

In Table~\ref{tab: mAP_table}, we report the mAP for the 2D object detectors listed above. We observe that the most widely used 2D object detectors, that perform well on the state-of-the-art autonomous driving datasets, like the Waymo Open Motion Dataset~\cite{waymo} and the KITTI dataset~\cite{kitti}, do not perform well on \rain. More specifically, the detectors achieve $37\%-65\%$ and $23\%-81\%$ mAP on the Waymo and KITTI datasets, respectively, while the same methods achieve $8\%-31\%$ mAP on the \rain dataset. In other words, the best possible result on \rain is $\frac{1}{2} \times$ and $\frac{1}{3} \times$ the best result on the Waymo and KITTI datasets, respectively. In Table~\ref{tab: swint_table}, we compare \rain in depth with the Waymo dataset using the Swin-T method~\cite{liu2021swin}, which is currently one of the top performing methods on the standard COCO 2D object detection benchmark leaderboard~\cite{coco}. The Swin-T method performs $14\%$ better on the Waymo Dataset.

There are two possible reasons for performance degradation on \rain. First, 2D detectors are typically pre-trained on MS COCO~\cite{coco} and ImageNet~\cite{deng2009imagenet}, which contain only up to $9$ categories of the commonly occurring traffic agents. This was not an issue for detectors on existing datasets like Waymo and KITTI since those datasets contain a subset of those $9$ classes. \rain, on the other hand, contains $16$ agent categories that are approximately equally distributed. The approximately $7-8$ traffic agent categories that are contained in \rain but do not appear in MS COCO are novel to these 2D object detectors and are not classified correctly. 

The other reason why object detection deteriorates on \rain is due to the challenging traffic environments in \rain. More specifically, \rain contains many challenging scenarios such as bad weather, nighttime traffic, rural area, high density traffic, etc. (see Figure~\ref{fig: rulebreak}). We analyze the effect of meta-features such as traffic conditions (density and heterogeneity), road conditions, weather, and time-of the day on 2D object detection and present this analysis in Table~\ref{tab: meta_od}. For this analysis, we form separate test sets corresponding to each label in a meta-feature (for example, we have two test sets for day and night). Most datasets contain videos of medium density traffic. In Table~\ref{tab: meta_od}, we see that the performance of the DETR, Deformable DETR, YOLOv3, and CenterNet suffers as the traffic density increases from medium to high. Similar reasoning can be made for other factors--object detection is less effective for homogeneous traffic, in rural areas, at nighttime, and in rainy weather. In most datasets, the number of annotated data samples with these adverse and challenging factors are a fraction of the entire dataset, which partly explains why 2D detectors are more successful on those datasets. The analysis in this section empirically validates the difficulty that the heavy-tail problem poses to perception tasks in autonomous driving. 

\subsection{Multi-Agent Behavior Recognition}
\label{subsec: exp_behavior_prediction}

Multi-agent behavior recognition (MABR) is the task of first localizing agents in a video followed by classifying their behaviors. This task has drawn attention in recent years and plays an important role in autonomous driving. Unlike object detection, which can be accomplished solely by observing visual appearances, MABR reasons about the actors’ interactions with the surrounding context, including environments, other people and objects.

\noindent\textbf{Dataset Preparation:} The \rain dataset is ideal for spatio-temporal MABR due to the availability of bounding box annotations and their corresponding behavior labels for more than $1231$ video clips, each lasting one minute in duration, and over $2$ million annotated frames. We use $1000$ video clips for training and $231$ video clips for testing. As the guidelines of the benchmarks, we evaluate 16 behavior classes with mean Average Precision (mAP) as the metric, using a frame-level IoU threshold of 0.5.

\begin{table}[t]
\centering
\caption{\textbf{ACAR-Net on AVA and \rain:} We applied currently the state-of-the-art multi-agent action recognition approach on AVA to our METEOR dataset. (PT: pre-train, BS: batch size, Opt.: Optimization, LR: learning rate, WD: weight decay, FR(RX-101): Faster R-CNN (ResNeXt-101), Kin.-700: Kinetics-700, CR(Swin-T): Cascade R-CNN (Swin-T))}  
\resizebox{\columnwidth}{!}{
\begin{tabular}{rccccccc}
\toprule[1.25pt]
Dataset & Detector & PT & BS & Opt. & LR & WD & mAP\\
\hline 
AVA~\cite{gu2018ava} & FR(RX-101) & Kin.-700  & $32$ & $SGD$ & $0.008$ & $1\mathrm{e}{-7}$ & $30.0$\\
\midrule
\textbf{\rain} & CR(Swin-T) &Kin.-700 & $32$ & $SGD$ & $0.008$ & $1\mathrm{e}{-7}$ & $\bm{6.10}$\\
\bottomrule[1.25pt]
\end{tabular}
}
\label{tab: acar_table}
\vspace{-15pt}
\end{table}

\noindent\textbf{Framework:} We use the ActorContext-Actor Relation Network (ACAR-Net)~\cite{pan2021actor} which builds upon a novel high-order relation reasoning operator and an actor-context feature bank for indirect relation reasoning for spatio-temporal action localization. This framework is composed of an object detector, backbone network, and ACAR components.

\noindent\textbf{Object Detector:} For the object detection step, we use the Swin-T detector, generated by combining a Cascade R-CNN~\cite{cai2018cascade} with a Swin-T~\cite{liu2021swin} backbone. The model is pre-trained on ImageNet and MS COCO, and fine-tuned on \rain using the same settings as Swin-T~\cite{liu2021swin}: multi-scale training~\cite{carion2020end} (resizing the input with the shorter side between $480$ and $800$ and the longer side at most $1333$), AdamW~\cite{loshchilov2017decoupled} optimizer (initial learning rate of $1\mathrm{e}{-4}$, weight decay of $0.05$, and batch size of $16$), and $1\times$ schedule ($12$ epochs).

\noindent\textbf{Backbone Network:} Following ACAR-Net~\cite{pan2021actor}, we use SlowFast networks~\cite{feichtenhofer2019slowfast} as the backbone in the localization framework and double the spatial resolution of res5. We conduct experiments using a SlowFast R-101 $8\times8$, pre-trained on the Kinetics-700 dataset~\cite{carreira2019short}, without non-local blocks. The inputs are 64-frame clips, where we sample $T = 8$ frames with a temporal stride $\tau= 8$ for the slow pathway, and $\alpha T$($\alpha = 4$) frames for the fast pathway. 

\noindent\textbf{Training Settings:} We train ACAR-Net using synchronous SGD with a batch size of $16$. For the first $3$ epochs, we use a base learning rate of $0.008$, which is then decreased by a factor of $10$ at iterations $4$ epochs and $5$ epochs. We use a weight decay of $1\mathrm{e}{-7}$ and Nesterov momentum of $0.9$. We use both ground-truth boxes and predicted object boxes  for training. For inference, we scale the shorter side of input frames to $384$ pixels and use detected object boxes with scores greater than $0.85$ for final behavior classification.

\noindent\textbf{Results:} We compare \rain with the AVA dataset~\cite{gu2018ava} as the latter is the state-of-the-art in multi-agent action recognition. In Table~\ref{tab: acar_table}, we show that the current state-of-the-art approach, ACAR, achieves $30.0\%$ mAP on AVA but yields $6.1\%$ mAP on \rain. There are several reasons why ACAR performs better on AVA. AVA focuses exclusively on only one target, humans, a category which most state-of-the-art object detectors can detect with ease. Furthermore, the videos in the AVA dataset consist of high-definition movies, in which agents (actors) are clearly visible, the background is simple, and the movements performed are also exaggerated and easier to identify. \rain, on the other hand consists of $16$ different categories of agents from  vehicles to animals, most of which are novel for most detectors and therefore hard to detect. Moreover, the movements of the agents on the road are very fast, making them hard to capture. Finally, different agents have different motion patterns; for example, pedestrians move differently than vehicles and buses move differently than motorbikes. All of these factors collectively contribute to the complexity of MABR in dense, heterogeneous, and unstructured traffic scenarios. Our experiments and analysis show that there is much room for improvement and our hope with \rain is that it provides the research community the resources it needs to tackle this important problem.

\section{Conclusion, Limitations and Future Work}

We present a new dataset, \rain, for autonomous driving applications in dense, heterogeneous, and unstructured traffic scenarios. rain~consists of more than $1000$ one-minute video clips, over $2$ million annotated frames with 2D and GPS trajectories for $16$ unique agent categories, and more than $13$ million bounding boxes for traffic agents. We found that current models for object detection and multi-agent behavior prediction fail on the \rain dataset. \rain marks the first step towards the development of more sophisticated and robust perception models for dense, heterogeneous, and unstructured scenarios.

Our dataset has some limitations. While \rain~contains bounding box information for the surrounding agents, we currently do not provide trajectory information from a fixed reference frame. One would have to use depth estimation techniques to extract such trajectories. Furthermore, our dataset does not contain HD maps ad pointcloud data, which are used in many applications.  For future work, we hope that our dataset can benefit in terms of design and evaluation of new motion forecasting and behavior prediction algorithms in dense and heterogeneous traffic. 
Finally, we hope to include semantic segmentation capability as part of \rain~by providing pixel labels for each object.

\bibliographystyle{IEEEtran}
\bibliography{refs}

\end{document}